\documentclass[letterpaper, 10 pt, conference]{ieeeconf}
\IEEEoverridecommandlockouts
\let\labelindent\relax
\usepackage{cite}
\usepackage{mathrsfs}
\usepackage{lipsum}
\usepackage{graphicx}
\usepackage{float}
\usepackage{amsfonts}
\usepackage{dsfont}
\usepackage{setspace}
\usepackage{algorithm}
\usepackage{algorithmicx}
\usepackage{algpseudocode}
\usepackage{threeparttable}

\usepackage{array}
\usepackage{soul}

\let\labelindent\relax
\usepackage[utf8]{inputenc} 
\usepackage[T1]{fontenc}    
\usepackage{graphicx} \graphicspath{ {figures/} }
\usepackage{amsmath,amssymb,mathabx,amsbsy,mathtools,etoolbox}
\usepackage{times}
\usepackage{graphicx}
\usepackage{amsmath,amssymb,mathbbol}
\usepackage{acronym}
\usepackage{enumitem}
\usepackage[breaklinks,colorlinks,linkcolor=black]{hyperref}
\usepackage{balance}
\usepackage{xspace}
\usepackage{setspace}
\usepackage[skip=3pt,font=small]{subcaption}
\usepackage[skip=3pt,font=small]{caption}
\usepackage[dvipsnames,svgnames,x11names]{xcolor}
\usepackage[capitalise,nameinlink]{cleveref}
\usepackage{booktabs,tabularx,colortbl,multirow,multicol,array,makecell}
\usepackage{dblfloatfix}
\usepackage[misc]{ifsym}
\usepackage{pifont}
\usepackage{cite}

\makeatletter
\DeclareRobustCommand\onedot{\futurelet\@let@token\@onedot}
\def\@onedot{\ifx\@let@token.\else.\null\fi\xspace}

\makeatother

\makeatletter
\def\BState{\State\hskip-\ALG@thistlm}
\makeatother

\makeatletter
\renewcommand{\paragraph}{%
  \@startsection{paragraph}{4}%
  {\z@}{0ex \@plus 0ex \@minus 0ex}{-1em}%
  {\hskip\parindent\normalfont\normalsize\bfseries}%
}
\makeatother

\crefname{algorithm}{Alg.}{Algs.}
\Crefname{algocf}{Algorithm}{Algorithms}
\crefname{section}{Sec.}{Secs.}
\Crefname{section}{Section}{Sections}
\crefname{table}{Tab.}{Tabs.}
\Crefname{table}{Table}{Tables}
\crefname{figure}{Fig.}{Fig.}

\definecolor{gblue}{HTML}{4285F4}
\definecolor{gred}{HTML}{DB4437}
\definecolor{ggreen}{HTML}{0F9D58}

\definecolor{mygray}{gray}{.92}

\begin{document}

\title{\LARGE \bf ASPIRe: An Informative Trajectory Planner with Mutual Information Approximation for Target Search and Tracking}

\author{Kangjie Zhou\textsuperscript{1}, 
Pengying Wu\textsuperscript{1},
Yao Su\textsuperscript{2},
Han Gao\textsuperscript{1},
Ji Ma\textsuperscript{1},
Hangxin Liu\textsuperscript{2},
and Chang Liu\textsuperscript{1}
\thanks{*This work was sponsored by Beijing Nova Program (20220484056) and the National Natural Science Foundation of China (62203018).}
\thanks{\textsuperscript{1}Department of Advanced Manufacturing and Robotics, College of Engineering, Peking University. \textsuperscript{2}National Key Laboratory of General Artificial Intelligence, Beijing Institute for General Artificial Intelligence (BIGAI).
All correspondences should be sent to Chang Liu.}}

\maketitle

\begin{abstract}
This paper proposes an informative trajectory planning approach, namely, \textit{adaptive particle filter tree with sigma point-based mutual information reward approximation} (ASPIRe), for mobile target search and tracking (SAT) in cluttered environments with limited sensing field of view. We develop a novel sigma point-based approximation to accurately estimate mutual information (MI) for general, non-Gaussian distributions utilizing particle representation of the belief state, while simultaneously maintaining high computational efficiency.
Building upon the MI approximation, we develop the Adaptive Particle Filter Tree (APFT) approach with MI as the reward, which features belief state tree nodes for informative trajectory planning in continuous state and measurement spaces. An adaptive criterion is proposed in APFT to adjust the planning horizon based on the expected information gain. Simulations and physical experiments demonstrate that ASPIRe achieves real-time computation and outperforms benchmark methods in terms of both search efficiency and estimation accuracy.
\end{abstract}

\setstretch{0.95}
\section{Introduction}
Target search and tracking (SAT) using autonomous robots play significant roles in various military and civilian applications such as surveillance~\cite{lozano2022surveillance}, disaster response~\cite{aggravi2021haptic}, and environment exploration~\cite{niroui2019deep}. 
In these applications, the robot first needs to explore the environment and search for the target. 
Once the target is detected, the robot enters the tracking stage to maintain the target inside the field of view (FOV). Mainstream strategies formulate SAT as an information-gathering problem, where the robot trajectory is optimized to obtain informative measurements to reduce the expected target state uncertainty~\cite{charrow2014approximate, hoffmann2009mobile}.
Solving this problem mainly encompasses target state estimation to handle inherent sensing uncertainty and motion planning to generate informative trajectories for more precise target localization.

State estimation is a crucial part in SAT, and the filtering approaches for state estimation in SAT primarily include the grid-based Bayesian filters 
\cite{furukawa2006recursive}, the Kalman filter variants~\cite{chung2006decentralized} and the particle filter~\cite{tisdale2008multiple}. 
Since the particle filter can form non-parametric representations for arbitrary probability distributions, the particle filter has shown superior performance in many SAT tasks when compared to its competitors~\cite{ryan2010particle, charrow2014approximate}.

\begin{figure}[t]
\centering
\includegraphics[width=\linewidth]{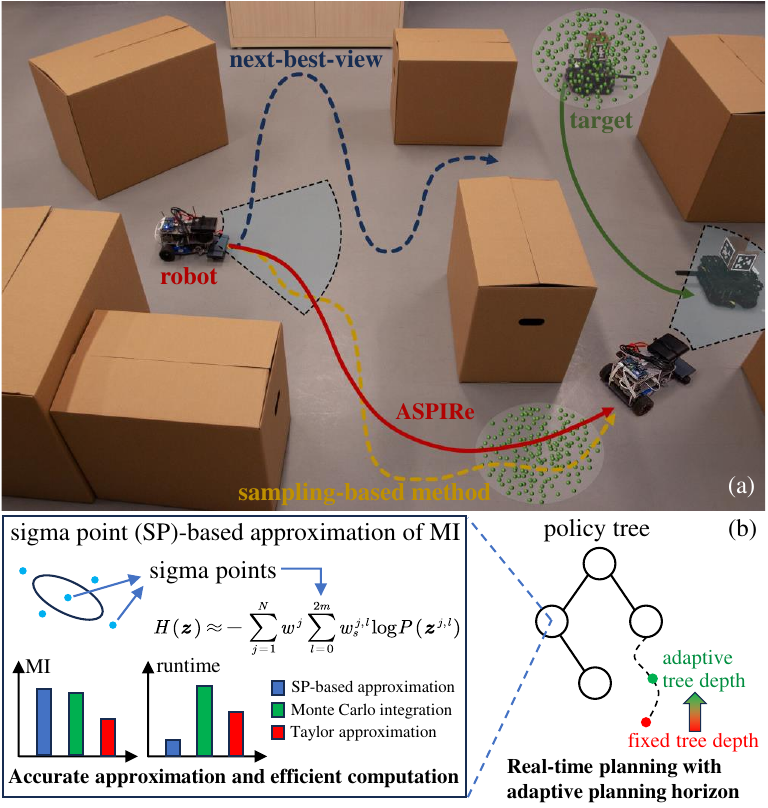}
\caption{(a) Comparison of different planning methods in mobile target SAT with prior uncertainty (green particles show the initial target state distribution). 
Suffering from the myopic horizon, the next-best-view method tends to randomly explore the environment. Although the sampling-based method can make long-term planning, ASPIRe generates a smoother and shorter trajectory.
(b) ASPIRe combines SP-based MI approximation with an adaptive planning horizon instead of a fixed one in the policy tree, enabling the maintenance of abundant particles for precise distribution representation, accurate reward approximation, and efficient planning.}
\label{fig:cover_picture}
\end{figure}

Several objective functions have been utilized for robot trajectory planning to facilitate SAT. 
While early works used the probability of detection as the objective function~\cite{hollinger2009efficient,bourgault2006optimal, tisdale2009autonomous}, recent studies have shown that information-theoretic objectives, 
especially the mutual information (MI)~\cite{asgharivaskasi2022active, yang2023learning}, demonstrate superior performance in encouraging robots to proactively gather target information.
However, calculating MI for non-Gaussian belief states in SAT, typically caused by nonlinear sensor models and target dynamics, involves integration over the continuous state and measurement spaces, lacking a general analytical expression. 
To address this computational challenge, the particle filter has been employed to sample the continuous state space, facilitating MI computation~\cite{ ryan2010particle, julian2012distributed}. However, integrating over the continuous measurement space remains a significant hurdle. 
To mitigate this issue, the approximation of MI based on Taylor expansion has been combined with the particle filter to approximate integration in the continuous measurement space \cite{charrow2014approximate}.
Nevertheless, the approximation accuracy is sacrificed for the purpose of computational efficiency, leading to non-trivial approximation error.

Various planning methods have been utilized for information gathering, employing simplification techniques to make the computation of information-theoretic objectives tractable. 
The greedy policy that chooses the next-best-view (NBV) has been widely adopted for its low computation complexity~\cite{julian2012distributed,bourgault2006optimal}, but such a strategy always falls short in complex environments due to its myopic nature. In response to this challenge, Jadidi et al.~\cite{ghaffari2019sampling} proposed sampling-based methods for non-myopic informative path planning, which enables online replanning by incorporating an automatic stopping criterion. 
Nonetheless, the discrete measurement space is used to simplify MI computation, which inevitably introduces discretization errors. Liu et al.~\cite{liu2017model} used model predictive control to generate informative trajectories for SAT in continuous spaces, yet relied on a restrictive Gaussianity assumption of belief state for computability, which severely limits the generality of the method.

In recent years, Monte Carlo Tree Search (MCTS) has gained popularity as an online approach for non-myopic planning due to its ability to allocate computation resources to more valuable subtrees to prevent exhaustive search and has been utilized in trajectory planning for SAT tasks~\cite{vanegas2016uav, goldhoorn2018searching, wandzel2019multi}. In order to handle non-Gaussian belief states, MCTS combines the particle filter for belief inference, and uses the progressive widening strategy to allow for long planning horizon in continuous spaces, as exemplified by PFT-DPW~\cite{sunberg2018online} and IPFT~\cite{fischer2020information}. However, due to computationally expensive belief inference, these methods usually employ a limited number of particles, which compromises estimation accuracy and cannot meet the needs of many SAT tasks, especially under multimodal prior uncertainty.

In this work, we propose the \underline{a}daptive particle filter tree with \underline{s}igma \underline{p}oint-based mutual \underline{i}nformation \underline{re}ward approximation (ASPIRe) to generate non-myopic informative trajectories for mobile target SAT under continuous state and measurement spaces in cluttered environments (\Cref{fig:cover_picture}). 

The main contributions can be summarized as follows:
\begin{itemize}
\item We propose a novel sigma point (SP)-based approximation approach to compute the predictive MI under continuous state and measurement spaces for non-parametric belief states, while taking the limited sensing FOV into account. The approximation is more accurate and computationally efficient than state of the arts.
\item We develop an adaptive particle filter tree (APFT) approach to generate kinematically feasible, informative trajectories. An adaptive criterion is proposed for automatic termination in tree construction to improve the search efficiency. 
\item We combine APFT with SP-based approximation to obtain ASPIRe that enables online replanning and accurate target localization and tracking with abundant particles. Simulations and physical experiments demonstrate that ASPIRe achieves superior real-time computational capability, search efficiency, and estimation accuracy.
\end{itemize}

\section{Problem Formulation}
\subsection{System Models \label{subsec:system-model}}
Consider a discrete-time kinematic model for the robot, 
\begin{equation}
\small
\boldsymbol{x}_{k+1}^{r}=\mathbf{f}^r(\boldsymbol{x}_{k}^{r},\boldsymbol{u}_{k}^{r}),
\label{eqn:1}
\end{equation}
where $\mathbf{f}^r$ is the kinematic model, $\boldsymbol{x}_{k}^{r}$ and $\boldsymbol{u}_{k}^{r}$ denote the robot state and control at time step $k$, respectively, and the superscript $r$ represents the robot.
The target state and control are denoted as $\boldsymbol{x}_{k}^{t}$ and $\boldsymbol{u}_{k}^{t}$, where the superscript $t$ represents the target.
The target motion model is defined as 
\begin{equation}
\small
\boldsymbol{x}_{k+1}^{t}=\mathbf{f^{\mathit{t}}}(\boldsymbol{x}_{k}^{t},\boldsymbol{u}_{k}^{t})+\boldsymbol{\eta}_{k},\boldsymbol{\eta}_{k}\sim\mathcal{N}(0,\mathbf{Q}),
\label{eqn:motion model}
\end{equation}
where $\mathbf{f}^t$ represents the kinematic model. Here $\boldsymbol{\eta}_k$ is a Gaussian noise with zero mean and covariance
matrix $\mathbf{Q}$.

Due to the limited sensing domain and obstacle occlusion, when the target is outside the FOV, the robot cannot detect the target and no measurement can be obtained. To reflect the intermittency of sensor measurements, a binary parameter $\gamma_{k}$ is defined to indicate if the target is inside the FOV ($\gamma_{k}=1$) or not ($\gamma_{k}=0$), and the measurement model is~\cite{gao2024probabilistic}:
\begin{equation}
\small
\boldsymbol{z}_{k}=\begin{cases}
\mathbf{h}(\boldsymbol{x}_{k}^{r},\boldsymbol{x}_{k}^{t})+\boldsymbol{\varepsilon}_{k},\boldsymbol{\varepsilon}_{k}\sim\mathcal{N}(0,\mathbf{\Sigma}) & \;\gamma_{k}=1\\
\varnothing & \;\gamma_{k}=0
\end{cases},
\label{eqn:sensing model}
\end{equation}
where $\boldsymbol{z}_{k}\in \mathbb{R}^{m}$ is the sensor measurement, $\mathbf{h}$ is the observation function, and $\boldsymbol{\varepsilon}_{k}$
is a zero-mean Gaussian white noise with covariance matrix $\mathbf{\Sigma} \in \mathbb{R}^{m\times m}$.

\subsection{Belief MDP Formulation with Particle Filter}
We formulate the SAT problem as a finite-horizon belief Markov decision process (MDP) $(h,\mathcal{B},\mathcal{A},\tau,\mathcal{R},\gamma)$, with belief state space $\mathcal{B}$, action space $\mathcal{A}$, belief transition model $\tau$, planning horizon $h$, discount factor $\gamma$, and reward function $\mathcal{R}$, which will be detailed in \Cref{sec:mutual-information}.
The robot state is assumed to be fully known, and the belief state is defined as $\boldsymbol{B}_k = [\boldsymbol{x}_{k}^r, P(\boldsymbol{x}_{k}^t)]\in \mathcal{B}$, where $P(\boldsymbol{x}_{k}^t)$ represents the probability distribution of the target state, denoted as the target belief state.
To handle potential nonlinearity in target dynamics and sensor models, especially due to the limited sensing domain, we use the particle filter to estimate the target belief because of its capability to represent arbitrary probability distributions.
Specifically, the target belief can be approximated by weighted particles as $P(\boldsymbol{x}_{k}^t) \approx\sum_{j=1}^{N}w_{k}^{j}\delta(\boldsymbol{x}_{k}^{t}-\tilde{\boldsymbol{x}}_{k}^{t,j})$, where $\tilde{\boldsymbol{x}}_{k}^{t,j}$ is the $j$th particle, $w_{k}^{j}$ is the corresponding weight, $N$ is the number of particles, and $\delta(\cdot)$ is a Dirac function. 
The action $\boldsymbol{a}_k = [\boldsymbol{u}_k^{r},\boldsymbol{u}_k^{t}]\in \mathcal{A}$ encompasses the control inputs of the robot and the target, respectively.
The belief transition model $\tau$ is defined as $\boldsymbol{B}_{k+1}=\tau(\boldsymbol{B}_k,\boldsymbol{a}_k,\boldsymbol{z}_k)=[\boldsymbol{x}_{k+1}^r, P(\boldsymbol{x}_{k+1}^t)]$,
where $\boldsymbol{x}_{k+1}^r$ is propagated based on \cref{eqn:1}, and ${b}_{k+1}^t$ is updated using particle filtering with the following prediction step (\cref{eqn:pf_pred}) and update step (\cref{eqn:pf_upd}),
\begin{equation}\label{eqn:pf_pred}
\small
\tilde{\boldsymbol{x}}_{k+1}^{t,j}\sim\mathcal{N}(\mathbf{f}^t(\tilde{\boldsymbol{x}}_{k}^{t,j},\boldsymbol{u}_{k}^t),\mathbf{Q}),\quad j=1,\ldots,N,
\end{equation}
\begin{equation}\label{eqn:pf_upd}
\small
w_{k+1}^{j}=\frac{P(\boldsymbol{z}_{k+1}|\tilde{\boldsymbol{x}}_{k+1}^{t,j})}{\sum_{j=1}^{N}P(\boldsymbol{z}_{k+1}|\tilde{\boldsymbol{x}}_{k+1}^{t,j})}w_{k}^{j},\quad j=1,\ldots,N,
\end{equation}
where the particles' states are first forward predicted based on the target dynamics, then the weights are updated with new measurements.
To alleviate particle degeneracy, a resampling procedure is performed subsequently. We use low variance resampling strategy to mitigate sampling error~\cite{thrun2005probabilistic}.

We aim to obtain the optimal policy $\boldsymbol{\pi}^{\ast}  = (\pi_1^{\ast},\ldots,\pi_h^{\ast})$ that maximizes the expected total discounted reward, 
\begin{equation}
\small
\boldsymbol{\pi}^{\ast}=\mathop{\arg\max}\limits_{\boldsymbol{\pi}} \mathbb{E} \left[\sum_{t=k}^{k+h-1} \gamma^{t-k}\mathcal{R}(\boldsymbol{B}_{t},\boldsymbol{a}_{t}) \bigg| \boldsymbol{a}_{t}=\pi_{t-k+1}(\boldsymbol{B}_{t}) \right],
\nonumber
\end{equation}
where $\mathbb{E}$ is the expectation over future beliefs. The robot then executes the optimal action $\boldsymbol{a}_{k}^{\ast}=\pi^{\ast}_{1}(\boldsymbol{B}_{k})$ and replans at the next time step based on the new measurements. 

\section{Sigma Point-Based Mutual Information Approximation \label{sec:mutual-information}}
\subsection{Reward Function Definition}
To gather more target information from future observations, we define the reward as the MI between the target belief and predicted measurements,
\begin{equation}
\small
\mathcal{R}(\boldsymbol{B}_{k},\boldsymbol{a}_{k}) = I(\boldsymbol{x}_{k+1}^t;\boldsymbol{z}_{k+1})
= H(\boldsymbol{z}_{k+1})-H(\boldsymbol{z}_{k+1}|\boldsymbol{x}_{k+1}^t),
\label{obj_def}
\end{equation}
where $I$ and $H$ denote the MI and the entropy, respectively. 

Using particle filter, future belief can be approximated as
\begin{equation}
\small
P(\boldsymbol{x}_{k+1}^t) \approx \sum\nolimits_{j=1}^{N}w_{k}^{j}\delta(\boldsymbol{x}_{k+1}^t-\tilde{\boldsymbol{x}}_{k+1}^{t,j}),
\label{eqn:10}
\end{equation}
where the particle state is predicted based on the target dynamics, while the weight remains unchanged since the future measurement is unknown. 

First, we calculate the conditional measurement entropy in \cref{obj_def}
with the particle expression,
\begin{equation}
\small
\begin{split}
H(\boldsymbol{z}_{k+1}|\boldsymbol{x}_{k+1}^t) 
&\approx \sum\nolimits_{j=1}^{N} P(\tilde{\boldsymbol{x}}_{k+1}^{t,j})H(\boldsymbol{z}_{k+1}|\boldsymbol{x}_{k+1}^t=\tilde{\boldsymbol{x}}_{k+1}^{t,j})\\
&\approx \sum\nolimits_{j=1}^{N} w_{k}^{j}H(\boldsymbol{z}_{k+1}|\tilde{\boldsymbol{x}}_{k+1}^{t,j}),
\nonumber
\end{split}
\end{equation}
where the first equality is derived from the definition of conditional entropy and the second equality is obtained by substituting $P(\tilde{\boldsymbol{x}}_{k+1}^{t,j})$ with \cref{eqn:10}.
According to the measurement model, the likelihood is 
\begin{equation}
\small
P(\boldsymbol{z}_{k+1}|\tilde{\boldsymbol{x}}_{k+1}^{t,j}) 
=  \begin{cases}
\mathcal{N}(\boldsymbol{z}_{k+1};\mathbf{h}(\boldsymbol{x}_{k+1}^{r},\tilde{\boldsymbol{x}}_{k+1}^{t,j}),\mathbf{\Sigma}) & \gamma_{k+1}^{j}=1\\
{\mathds{1}}_{\boldsymbol{z}_{k+1}=\varnothing} & \gamma_{k+1}^{j}=0
\nonumber
\end{cases},
\end{equation}
where $\gamma_{k+1}^{j}$ denotes whether the $j$th particle is inside the FOV. If the $j$th particle can be detected, i.e., $\gamma_{k+1}^{j}=1$, the likelihood is an $m$-dimensional Gaussian distribution and its entropy has explicit expression $H_0 = \frac{m}{2}(\log2\pi+1)+\frac{1}{2}\log|\mathbf{\Sigma}|$.
Otherwise, no observation is expected to be obtained, and the corresponding entropy is zero. So we can derive that 
\begin{equation}
\small
H(\boldsymbol{z}_{k+1}|\tilde{\boldsymbol{x}}_{k+1}^{t,j})={\mathds{1}}_{\gamma_{k+1}^{j}=1} H_0.
\end{equation}

Next, we consider computing entropy $H(\boldsymbol{z}_{k+1})$. Utilizing the particle representation of target state distribution,
the measurement distribution $P(\boldsymbol{z}_{k+1})$ can be computed as
\begin{equation}
\small
\begin{split}
P(\boldsymbol{z}_{k+1}) & = \int P(\boldsymbol{z}_{k+1}|\boldsymbol{x}_{k+1}^t)P(\boldsymbol{x}_{k+1}^t)d\boldsymbol{x}_{k+1}^t \\
& \approx \sum\nolimits_{j=1}^{N}w_{k}^{j}P(\boldsymbol{z}_{k+1}|\tilde{\boldsymbol{x}}_{k+1}^{t,j}).
\end{split}
\end{equation}

When the $j$-th particle is in the FOV,  $P(\boldsymbol{z}_{k+1}|\tilde{\boldsymbol{x}}_{k+1|k}^{t,j})$ is a Gaussian distribution by the definition of the measurement model, and $P(\boldsymbol{z}_{k+1})$ follows a Gaussian Mixture Model (GMM), whose entropy has no closed form. 
Though the entropy of GMM can be numerically evaluated using Monte Carlo integration, the amount of samples required for accurate approximation usually results in large computational overhead and therefore limits this method in practice. 
The Taylor expansion has been used to approximate the entropy of GMM~\cite{huber2008entropy}, yet this method usually leads to a non-trivial approximation error. 
To overcome the undesirable tradeoff between the computational efficiency and the approximation accuracy in existing approaches, we propose to utilize the sigma points associated with each Gaussian component in GMM to approximate the GMM entropy, which will be detailed in the next subsection.

\subsection{Sigma Point-Based Entropy Approximation}

The entropy $H(\boldsymbol{z}_{k+1})$ can be computed as
\begin{equation}
\small
H(\boldsymbol{z}_{k+1})
=-\int P(\boldsymbol{z}_{k+1})\log P(\boldsymbol{z}_{k+1})d \boldsymbol{z}_{k+1} \approx -\sum_{j=1}^{N}w_{k}^{j} p_j,
\nonumber
\end{equation}
where $p_j = 
\displaystyle
\int P(\boldsymbol{z}_{k+1}|\tilde{\boldsymbol{x}}_{k+1}^{t,j})\log P(\boldsymbol{z}_{k+1})d\boldsymbol{z}_{k+1}$.
Denote $A$ as the set of indices that particles are inside the FOV,
i.e., $\forall i\in A, \gamma_{k+1}^{i}=1$. 
If $j\notin A$, then $p_j$ has an explicit expression,
\begin{equation}\label{eqn:sp_outFOV}
\small
\begin{split}
p_j = P(\varnothing|\tilde{\boldsymbol{x}}_{k+1}^{j})\log P(\varnothing) = \log \sum\nolimits_{i\notin A} w_{k}^{i}.
\end{split}
\end{equation}

If $j\in A$, we propose to utilize sigma points from the Unscented Transform~\cite{van2004sigma} to approximate $p_j$. Denote the sigma points and their weights corresponding to the $j$th Gaussian component as $[\tilde{\boldsymbol{z}}_{k+1}^{j,0},\ldots,\tilde{\boldsymbol{z}}_{k+1}^{j,2m}]^{T}$ and $\left[w_{s}^{j,0}, \ldots,w_{s}^{j,2m}\right]^{T},\forall j=1,\ldots,N$, respectively, and they are computed as follows~\cite{van2004sigma},
\begin{small}
\begin{equation}
\begin{split}
\tilde{\boldsymbol{z}}_{k+1}^{j,0} & =  \boldsymbol{\mu}_{j},\\
\tilde{\boldsymbol{z}}_{k+1}^{j,l} & =  \boldsymbol{\mu}_{j}+\big(\sqrt{(\lambda+m)\mathbf{\Sigma}}\big)_{l}, \quad\quad\quad l=1,\ldots,m\\
\tilde{\boldsymbol{z}}_{k+1}^{j,l} & =  \boldsymbol{\mu}_{j}-\big(\sqrt{(\lambda+m)\mathbf{\Sigma}}\big)_{l-m}, \,\,\,\,\quad l=m+1,\ldots,2m\\
w_{s}^{j,0} & = \frac{\lambda}{\lambda+m},\;w_{s}^{j,l} =  \frac{1}{2(\lambda+m)}, \quad l=1,\ldots,2m
\nonumber
\end{split}
\end{equation}
\end{small}

\noindent where $\lambda$ is a parameter that determines the sigma points spread, $\boldsymbol{\mu}_{j}=\mathbf{h}(\boldsymbol{x}_{k+1}^{r},\tilde{\boldsymbol{x}}_{k+1}^{t,j})$ is the mean of $P(\boldsymbol{z}_{k+1}|\tilde{\boldsymbol{x}}_{k+1}^{t,j})$, and $\big(\sqrt{(\lambda+m)\mathbf{\Sigma}}\big)_{l}$ is the $l$-th column of the matrix square root. 
Since the observation is $m$-d, there are $2m+1$ sigma points for each Gaussian component. Thus the $j$th Gaussian component can be approximated as
\begin{equation}
\small
P(\boldsymbol{z}_{k+1}|\tilde{\boldsymbol{x}}_{k+1}^{t,j})\approx\sum\nolimits_{l=0}^{2m}w_{s}^{j,l}\delta(\boldsymbol{z}_{k+1}-\tilde{\boldsymbol{z}}_{k+1}^{j,l}),
\end{equation}
and $p_j$ can be approximated as follows,
\begin{equation} \label{eqn:sp_inFOV}
\small
\begin{split}
p_j
& \approx \sum\nolimits_{l=0}^{2m}w_{s}^{j,l}\log P(\tilde{\boldsymbol{z}}_{k+1}^{j,l})\\
& \approx \sum\nolimits_{l=0}^{2m}w_{s}^{j,l}\log\sum\nolimits_{i\in A}  w_{k}^{i}P(\tilde{\boldsymbol{z}}_{k+1}^{j,l}|\tilde{\boldsymbol{x}}_{k+1}^{t,i}).
\end{split}
\end{equation}
By employing this approach, with \cref{eqn:sp_outFOV} and \cref{eqn:sp_inFOV}, we obtain an explicit expression to approximate the entropy $H(\boldsymbol{z}_{k+1})$.

\section{Adaptive Particle Filter Tree for Planning}

We propose the APFT, a MCTS-based planning algorithm to generate informative trajectories for SAT tasks. 
By exploring possible sequences of actions and observations in the planning horizon, the proposed method constructs an asymmetric policy tree to direct the robot's actions.

\begin{algorithm}[!t]
\caption{\textbf{Adaptive Particle Filter Tree}}
\label{alg1}
\begin{algorithmic}[1]
\fontsize{8pt}{8pt}\selectfont
\State \textbf{function} \textsc{APFT}($\boldsymbol{B}_k$)
\State $n_r = \left \langle \varnothing,\varnothing,0,0 \right \rangle$
\For{$i\in1:n$}
    \State  \textsc{Simulate}($\boldsymbol{B}_k,n_r,h$) 
\EndFor
\State $n^{\ast} = \mathop{\arg\max}\limits_{n\in C(n_r)}$ \textsc{UCB}$(n)$
\State $\boldsymbol{a}^{\ast}$ = $\mathcal{H}(n^{\ast})$
\State \Return $\boldsymbol{a}^{\ast}$
\State \textbf{function} \textsc{Simulate}($\boldsymbol{B},n,d$)
\If{$d=0$} 
    \State $R=0$
\Else 
    \State $n_a,\boldsymbol{a} \leftarrow$ \textsc{SelectAction}($n$)
    \State $r \leftarrow$ $\mathcal{R}(\boldsymbol{B},\boldsymbol{a})$
    \State $\boldsymbol{B} \leftarrow \tau(\boldsymbol{B},\boldsymbol{a},\varnothing)$
    
    \If{$|C(n_a)| \leq k_o W(n_a)^{\alpha_o}$} 
        \State $\boldsymbol{z} \leftarrow $ \textsc{SampleNewObservation}($\boldsymbol{B}$)
        \State \textsc{AddNode}($\boldsymbol{z}$) 
        \State $\boldsymbol{B} \leftarrow $ $\tau$($\boldsymbol{B},\varnothing,\boldsymbol{z}$)
        \State $R \leftarrow r+\gamma \cdot$ \textsc{Rollout}($\boldsymbol{B},d-1$)
    \Else
        \State $n_o,\boldsymbol{z} \leftarrow$ \textsc{SelectObservation}($n_a$)
        \State $\boldsymbol{B} \leftarrow $ $\tau$($\boldsymbol{B},\varnothing,\boldsymbol{z}$)
        \State $R \leftarrow r+\gamma \cdot$ \textsc{Simulate}($\boldsymbol{B},n_o,d-1$)
    \EndIf
    \State $W(n_c) \leftarrow W(n_c)+1$
    \State $W(n_a) \leftarrow W(n_a)+1$
    \State $Q(n_a) \leftarrow Q(n_a)+\frac{R-Q(n_a)}{W(n_a)}$
\EndIf
\State \Return $R$
\end{algorithmic}
\end{algorithm}

\subsection{Algorithm Overview}
The entire procedure of APFT is shown in \Cref{alg1}.
The algorithm proceeds with the current belief state $\boldsymbol{B}_k $ as input and the optimal action $\boldsymbol{a}^{\ast}$ as output. 
The algorithm iterates through simulating action-observation sequences to expand action nodes $n_a$ and observation nodes $n_o$ in the rewarding subtree, based on upper confidence bound (UCB) criterion~\cite{kocsis2006bandit}.
Each node $n=\left \langle \mathcal{H},C,W,Q \right \rangle$ includes action-observation history $\mathcal{H}$, children set $C$, visit count $W$, and estimated value $Q$.
The planning horizon and remaining tree depth are denoted by $h$ and $d$, respectively.

First, the root node $n_r$ is initialized (Line 2), and \textsc{Simulate} is called repeatedly to construct the policy tree.
Specifically, In \textsc{Simulate}, an action node is first selected by the UCB criterion (Line 13) from the robot action space, which consists of motion primitives generated by the robot kinematics to allow performing smooth trajectories while avoiding obstacles.
Then the information reward of taking action $\boldsymbol{a}$ in belief $\boldsymbol{B}$ is calculated, and the belief state is updated (Line 14-15).
Given the selected action node, an observation node is obtained by either sampling a new observation from the updated belief or selecting an existing observation node depending on the parameters $k_o$ and $\alpha_o$.
If a new observation is generated, it is inserted into the policy tree as a new child node, and \textsc{Rollout} is subsequently performed to estimate the accumulated reward $R$ of the new node (Line 17-20), which will be detailed in the next subsection. 
If, on the other hand, a previous observation node is chosen, \textsc{Simulate} is called recursively on the new belief and observation node (Line 22-24). 
At last, the information of visited nodes is updated (Line 26-28).
After performing the desired number of iterations, the optimal action is chosen from the children of the root node that maximizes the UCB.

\textcolor{blue}{}

\subsection{Adaptive Criterion for Efficient Tree Search}

\begin{algorithm}[!t]
\caption{\textbf{Adaptive Rollout}}
\label{alg2}
\begin{algorithmic}[1]
\fontsize{8pt}{8pt}\selectfont
\State \textbf{function} \textsc{Rollout}($\boldsymbol{B},d$)
\If{$d=0$} 
    \State $R=0$
\Else 
    \State $\boldsymbol{a} \leftarrow$ DefaultPolicy()
    \State $r \leftarrow$ $\mathcal{R}(\boldsymbol{B},\boldsymbol{a})$
    \State $\boldsymbol{B} \leftarrow$ $\tau$($\boldsymbol{B},\boldsymbol{a},\varnothing$)
    \If{$r>\delta_r$} 
        \State \Return $r$
    \Else 
        \State $R \leftarrow r+\gamma \cdot$ \textsc{Rollout}($\boldsymbol{B},d-1$)
    \EndIf
\EndIf
\State \Return $R$
\end{algorithmic}
\end{algorithm}

The standard tree search method typically employs a fixed planning horizon, which may be inefficient for target search tasks.
Concretely, if the fixed planning horizon is short, the limited search space makes it difficult to find a remote target, while setting a lengthy horizon will result in computational redundancy when the robot is close to the target.

To alleviate this limitation, we provide an adaptive termination criterion to dynamically adjust the length of the planning horizon to reach a desirable trade-off between efficiency and effectiveness.
As shown in \Cref{alg2}, when \textsc{Rollout} progresses recursively, if the reward of one iteration in rollout exceeds the threshold $\delta_{r}$ (Line 8), \textsc{Rollout} will terminate in advance rather than reach the defined maximum depth. 
This is because the increasing information gain demonstrates that the robot will find the target in future rollout steps, which is informative for target search, and thus the planning horizon can be decreased to reduce computation time. Benefitting from the computational efficiency achieved through the adaptive termination criterion and MI approximation method, we can adopt a large set of particles in the tree search to ensure accurate belief representation, while maintaining the ability to perform online operations.

\setstretch{0.96}
\section{Simulation}
We conduct simulations to validate the proposed method in MATLAB using a desktop with Intel Core i7 CPU@2.10GHz and 16GB RAM. The environment is set as a $50m\times50m$ planar space comprising multiple obstacles, and the map is known to the robot. 
We consider the robot state $\boldsymbol{x}_{k}^{r}=[x_{k}^{r},y_{k}^{r},\theta_{k}^{r}]^{T}\in \mathbb{R}^3$ includes the $x$-$y$ position $x_{k}^{r},y_{k}^{r}$ and orientation $\theta_{k}^{r}$ of the robot, and the control input $\boldsymbol{u}_{k}^r\in \mathbb{R}^2$ consists of the linear velocity $v_k^r$ and angular velocity $w_k^r$.
The robot motion models use the following unicycle model,
\begin{equation}
\small
\mathbf{f}^r(\boldsymbol{x}_{k}^r,\boldsymbol{u}_{k}^r)=\boldsymbol{x}_{k}^r+[v_{k}^r\cos\theta_{k}^r, v_{k}^r\sin\theta_{k}^r, w_{k}^r]^T \cdot \Delta t,
\end{equation}
where $\Delta t$ is the sampling interval. 
The target adopts the similar state, control input, and unicycle model as the robot, and its control is assumed to be known for the robot in simulations. 
For the sensing module, we adopt a range-bearing sensor with fan-shaped FOV, whose sensing range is from $1m$ to $6m$ and sensing angle is $90^{\circ}$. 
To demonstrate the generalization capability, we test the proposed method in $50$ scenarios where target trajectories are randomly generated.
\subsection{Approximation Accuracy and Efficiency}
\label{subsec:sigma-point}

To investigate the performance of SP-based approximation, we manually control the robot to follow the target, ensuring the target remains in the FOV, and compare MI approximated by different methods, including Monte Carlo integration, Taylor approximation~\cite{huber2008entropy} and SP-based approximation, referred as \textit{SP} for simplicity.
We also combine SP with the particle simplification method~\cite{charrow2014approximate}, referred to as \textit{SP with particle simplification}, which partitions the state space and generates a simplified particle set by replacing particles in the same cell with their weighted average to improve computational efficiency.

The covariance matrix in \cref{eqn:motion model} and \cref{eqn:sensing model} are $\mathbf{Q}=diag(0.5,0.5,0.1)$ and $\mathbf{\Sigma}=diag(0.5,0.05)$, respectively. 
Define the MI computed by Monte Carlo integration as $\mathcal{I}$, which is treated as the ground-truth value, and the MI computed by other methods as $\mathcal{I}_m$.
We record the absolute error $\varepsilon_a$, defined as $\varepsilon_a=|\mathcal{I}-\mathcal{I}_m|$, the relative error $\varepsilon_r$, defined as $\varepsilon_r=\frac{|\mathcal{I}-\mathcal{I}_m|}{\mathcal{I}}$, and the computational time $\tau$. 

\Cref{Table 1} shows the average results over $50$ scenarios. 
It indicates that SP obtains the minimal approximation error among all methods.
Besides, SP with particle simplification achieves the smallest computational time and is more accurate than 2nd Taylor approximation approach, indicating desirable advantages in both computational efficiency and approximation accuracy over the state of the arts.

\begin{table}[!t]
\caption{\textbf{Comparison of MI approximation methods}}
\begin{center}
\resizebox{0.85\linewidth}{!}{
\begin{threeparttable}
\begin{tabular}{|c|c|c|c|}
\hline 
Methods & $\varepsilon_a$ & $\varepsilon_r(\%)$ & $\tau (s)$\tabularnewline
\hline 
Monte Carlo integration & - & - & $0.0614$\tabularnewline
\hline 
0th Taylor Approximation & $0.4220$ & $38.9$ & $0.0070$\tabularnewline
\hline 
2nd Taylor Approximation & $0.0993$ & $8.66$ & $0.3502$\tabularnewline
\hline 
SP & $\mathbf{0.0395}$ & $\mathbf{3.42}$ & $0.0302$\tabularnewline
\hline 
SP with particle simplification & $0.0533$ & $4.69$ & $\mathbf{0.0017}$\tabularnewline
\hline 
\end{tabular}         
\end{threeparttable}}
\label{Table 1}
\end{center}
\end{table}

\subsection{Benchmark Comparisons}
\label{subsec:APFT_performance}

We compare ASPIRe with other benchmark methods to assess its performance in mobile target search and tracking. Concretely, we consider the next-best-view (NBV) strategy and IIG-tree approach~\cite{ghaffari2019sampling}, a sampling-based information gathering algorithm, as baselines to make comparisons. 
All methods utilize the same motion model and control constraints, and use SP-based MI approximation with particle simplification technique~\cite{charrow2014approximate} as the objective function.
Since the IIG-tree was proposed to solve the exploration problem originally, to adapt it to the tracking task, we adjust its sample policy from sampling the whole workspace to sampling the area nearby the robot once the target has been detected. 

\begin{figure}[!t] 
\centering
\includegraphics[width=0.45\textwidth]{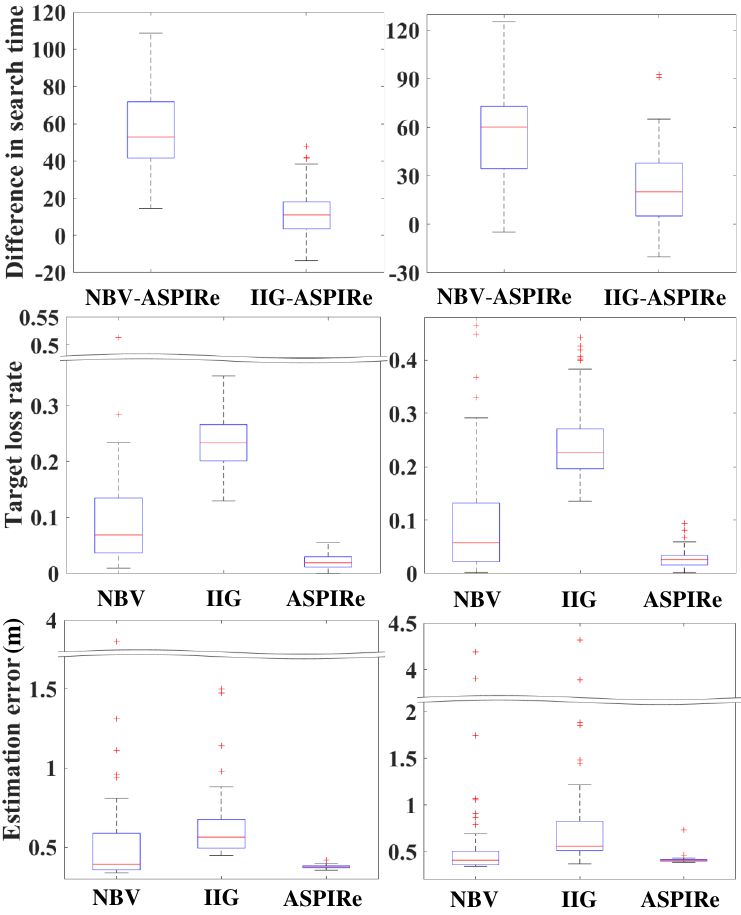} 
\caption{\textbf{Quantitative comparisons in the unimodal (left column) and multimodal case (right column).} NBV-ASPIRe and IIG-ASPIRe in the first row represent the difference in search time using the NBV strategy and IIG-tree, respectively, compared to ASPIRe.}
\label{fig:APFT}
\end{figure}

\begin{figure*}[!t] 
\centering
\includegraphics[width=0.95\textwidth]{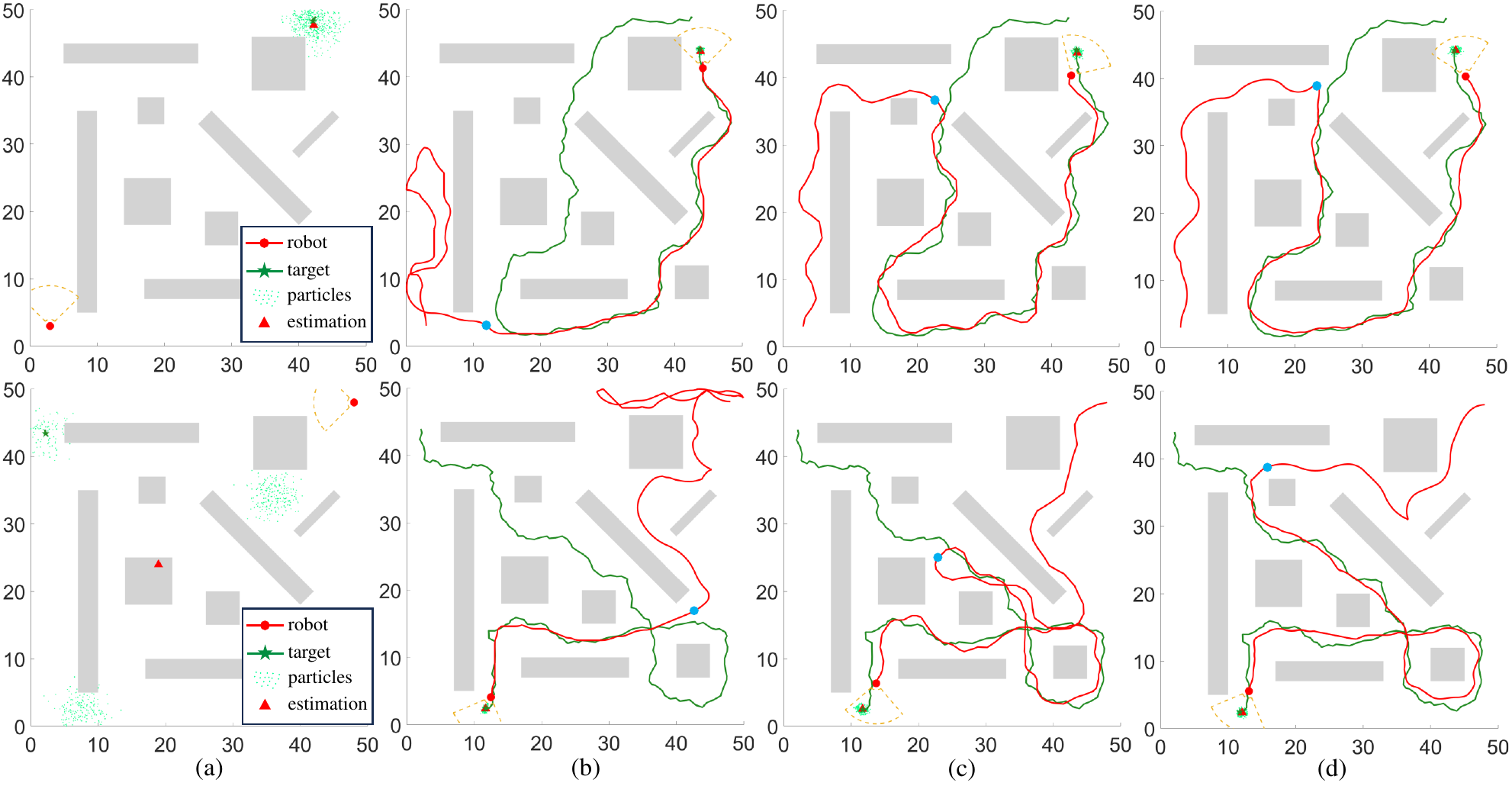}
\caption{\textbf{Qualitative comparisons in the unimodal case (top row) and multimodal case (bottom row).} (a) Initialization. (b) The NBV policy. (c) IIG-tree. (d) ASPIRe. The blue circle represents the moment when the target is detected. ASPIRe shows shorter search time and more stable tracking performance with smoother trajectories, even under distracting prior information. }
\label{Fig:unimodal_case} 
\end{figure*}

Simulations include the unimodal and multimodal case. 
The unimodal case is initialized with a Gaussian prior distribution whose mean is the target true pose, while the multimodal case considers a GMM with a Gaussian component initialized as above and two additional Gaussian distributions as distraction. 
The mean of disruptive Gaussian components are randomized in the obstacle-free space, and the covariance matrix for all components is $\mathbf{V}=diag(3,3,0.01)$. 
The weight assigned to the component with the target pose as mean is 0.2, while the disruptive components each have a weight of 0.4.
The covariance matrix are $\mathbf{Q}=diag(0.5,0.5,0.1)$ and $\mathbf{\Sigma}=diag(0.1,0.01)$.
Three metrics are evaluated for quantitative comparisons: the search time $t_{s}$, target loss rate $r_{los}$, and estimation error $\varepsilon_{est}$, defined as $\varepsilon_{est}=\frac{1}{T_{tra}} \sum_{k=1}^{T_{tra}} ||\boldsymbol{x}_{k}^{t}-\hat{\boldsymbol{x}}_{k}^{t}||$, where $T_{tra}$ is the total tracking time and $\hat{\boldsymbol{x}}_{k}^{t}$ is the average of the particles' positions. 
For each scenario, we repeat $5$ trials and average the results, which are shown in \Cref{fig:APFT}. 
\Cref{Fig:unimodal_case} show the qualitative comparisons in the unimodal and multimodal case.

\subsubsection{Unimodal Case}
The left column of \Cref{fig:APFT} shows the comparison results in the unimodal case.
Suffering from the myopic horizon, the NBV policy initially cannot gain information reward and can only randomly act to search for the target, which results in poor performance in target search.
IIG-tree can avoid obstacles and find the target faster than the NBV policy.
However, due to the random nature of samples, IIG-tree usually generates sinuous trajectories and loses sight of the target, leading to the highest average loss rate and estimation error.
ASPIRe significantly outperforms other methods by a large margin with less search time, considerably lower loss rate and localization error.
In $41$ out of $50$ scenarios, ASPIRe spends the least time to find the target. 
Compared to the NBV policy and IIG-tree, ASPIRe achieves $3$ times and $10$ times improvements in the target loss rate and yields $50\%$ and $80\%$ improvements in the estimation error, respectively. 
The quantitative comparisons show the effectiveness of ASPIRe in the unimodal case.

\subsubsection{Multimodal Case}
As depicted in the right column of \Cref{fig:APFT}, consistent with the unimodal case, ASPIRe maintains its advantages in search efficiency and tracking stability compared to other methods. 
The times of ASPIRe spends fewer simulation steps to find the target than the NBV policy and IIG-tree are $50$ and $40$ out of $50$ scenarios, and the proposed approach outperforms the benchmark methods by reducing the target loss rate and the estimation error at least $50\%$ and $60\%$, respectively.
The comparisons demonstrate the effectiveness and robustness of ASPIRe with unreliable prior information.

\section{Experiment}

\begin{figure}[!b] 
\centering
\includegraphics[width=0.45\textwidth]{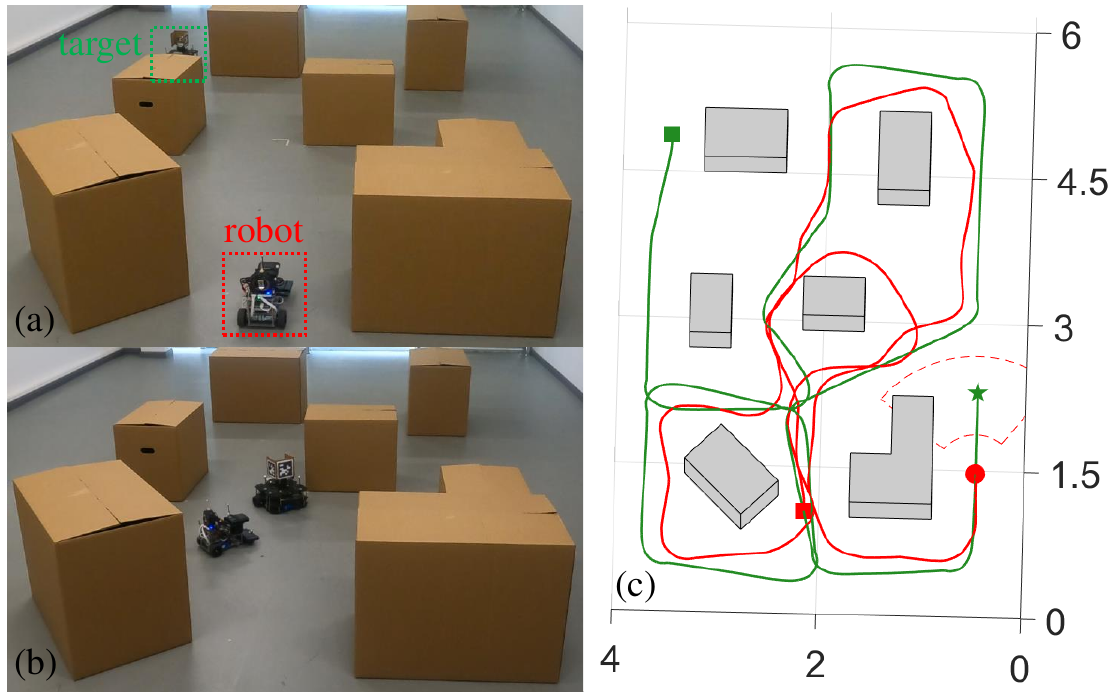}
\caption{\textbf{Indoor experiments with inaccurate prior information.} The red circle and green star show the current positions of the robot and the target, and the squares represent their starting positions. }
\label{Fig:exp}
\end{figure}

We investigate the performance of ASPIRe in real-world scenarios. The environment is a $6m \times 4m$ planar space and contains several obstacles. 
We use a Wheeltec ground robot to search for and track the moving target, which is a Turtlebot3 that carries three Apriltags~\cite{olson2011apriltag} for camera detection. Since the target control inputs are unavailable to the robot in real-world scenarios, we assume the target motion model follows an autonomous Markov model, 
\begin{equation}
\setlength{\abovedisplayskip}{3pt}
\setlength{\belowdisplayskip}{3pt}
\begin{split}
\boldsymbol{x}_{k+1}^{t}=\mathbf{f^{\mathit{t}}}(\boldsymbol{x}_{k}^{t})+\eta_{k},\eta_{k}\sim\mathcal{N}(0,\mathbf{Q}),
\end{split}
\end{equation}
and the robot has access to the model information. The motion noise is set as
$\mathbf{Q}=diag(0.05,0.05,0.01)$. 
We adopt a Vicon motion-capture system to measure the poses of the robot and target, which are treated as the ground-truth. 

We conducted three experiments in the multimodal case in the same environment. 
The first two scenarios have two Gaussian components as the initial belief, and the last scenario has three. 
For each scenario, we record the visibility rate $r_{vis}$ that denotes the percentage of time the target is in FOV once detected, the estimation error $\varepsilon_{est}$ as defined in \Cref{subsec:APFT_performance}, and the computational time $t_{run}$.
The robot successfully accomplishes the SAT task in all scenarios, and \Cref{Fig:exp} illustrates the trajectory performed by the robot in one scenario. 
As shown in \Cref{Table:exp}, the proposed method can achieve a considerably high visibility rate and low estimation error while maintaining 10.8Hz computing speed for real-time operation, which demonstrates the effectiveness of ASPIRe.
\begin{table}[!t]
\small
\caption{\textbf{Performance of ASPIRe in real-world experiments}}
\centering
\resizebox{0.8\linewidth}{!}{
\begin{tabular}{|c|c|c|c|}
\hline 
 & $r_{vis}$ ($\%$) & $\varepsilon_{est}$ ($m$) & $t_{run}$ ($s/step$)\tabularnewline
\hline 
Scenario 1 & $96.44$ & $0.0770$ & $0.0920$\tabularnewline
\hline 
Scenario 2 & $95.11$ & $0.0843$ & $0.0923$\tabularnewline
\hline 
Scenario 3 & $90.20$ & $0.0927$ & $0.0932$\tabularnewline
\hline 
\end{tabular}} \label{Table:exp}
\end{table}

\section{Conclusion}
This work presents ASPIRe, an informative trajectory planning approach for mobile target SAT in cluttered environments with limited sensing FOV. 
A novel sigma point-based approximation is proposed to accurately and efficiently compute mutual information in continuous measurement spaces.
APFT is also developed to generate informative trajectories while simultaneously achieving accurate target state estimation and efficient planning.
Simulation results demonstrate the superiority of ASPIRe compared to benchmark methods in terms of MI approximation, search efficiency and estimation accuracy. 
We also demonstrate the robustness and real-time performance of ASPIRe in real-world scenarios. 

\paragraph*{Acknowledgement}
We thank Dr. Meng Wang at BIGAI for his help with experiment photography.

\clearpage
{
\bibliographystyle{ieeetr}
\bibliography{ref}
}
\end{document}